\documentclass[runningheads]{llncs}
\usepackage[round,authoryear]{natbib}
\usepackage[T1]{fontenc}
\usepackage{amsmath, amssymb}
\usepackage{algorithm}
\usepackage{algpseudocode}
\usepackage{graphicx}
\usepackage{booktabs}
\usepackage{array}
\usepackage{float}
\usepackage{microtype}
\usepackage[misc]{ifsym}

\usepackage[table]{xcolor}
\usepackage{multirow}
\usepackage{hyperref} 
\usepackage{mwe}
\usepackage{tikz}
\usetikzlibrary{positioning,arrows.meta}

\begin{document}

\title{Trustworthy Calibration or Vulnerability? Unveiling the Over-Calibration Paradox and Attack Surface in Rashomon Sets}
\titlerunning{Trustworthy Calibration or Vulnerability?}

\author{Mustafa Cavus\orcidID{0000-0002-6172-5449}}
\authorrunning{M. Cavus}
\institute{Eskisehir Technical University, Department of Statistics, 26555 Eskisehir, Turkiye \\
\email{mustafacavus@eskisehir.edu.tr}}

\maketitle
\begin{abstract}
Post-hoc calibration is standard practice for aligning classifier confidence scores with empirical frequencies in high-stakes decision systems. However, its downstream effects on decision stability across competing, equally accurate models remain largely unexamined. In this paper, we formalize the \textit{over-calibration paradox} within the framework of \textit{predictive multiplicity}, reflecting the Rashomon effect where competing models with near-identical loss assign conflicting predictions to the same individual. Evaluating nine credit risk portfolios, we examine the impact of Platt Scaling, Isotonic Regression, and Beta Calibration on individual prediction agreement and expected calibration error. We identify two distinct operating regimes. In systematically miscalibrated models, post-hoc calibration consistently aligns probability estimates and reduces prediction disagreement. Conversely, when baseline models are already well-calibrated, post-hoc calibration, particularly non-parametric estimators constrained by finite sample sizes, overfits to hold-out noise. This distortion shifts decision thresholds near the classification boundary, paradoxically increasing both expected calibration error and predictive multiplicity. From an auditing and security standpoint, this paradox exposes a practical vulnerability to soft adversarial manipulation, whereby an interested party or automated model search can exploit calibration choices to steer specific individual outcomes while preserving aggregate performance and passing black-box evaluations. 

\end{abstract}

\section{Introduction}

Machine learning models are increasingly deployed in high-stakes decision-making domains such as healthcare, finance, and public policy, where individual-level trustworthiness is as critical as model-level prediction performance \citep{doshi2017}. In these sensitive pipelines, systems are expected to satisfy two fundamental pillars of trustworthy AI simultaneously: \textit{probabilistic reliability}--ensuring predicted probabilities reflect empirical real-world frequencies and \textit{prediction stability}--ensuring that the final decision does not depend on an arbitrary choice among equally accurate models \citep{decker2025}. While a massive body of literature addresses these pillars independently under the umbrellas of \textit{classification calibration} \citep{niculescu2005,guo2017} and \textit{predictive multiplicity} \citep{marx2020}, their structural interplay remains largely unexamined.

The phenomenon of \textit{predictive multiplicity} is deeply rooted in the \textit{Rashomon effect} \citep{breiman2001}, which characterizes the existence of a large set of near-optimal models consistent with the observed data. When competing models achieve indistinguishable overall performance yet assign conflicting outcomes to the same instances, automated decision pipelines subject individuals to unjustifiable outcome variability \citep{frohnapfel2026}. Such decision instability is particularly acute for vulnerable or minority subpopulations, which often bear a disproportionate burden of predictive multiplicity \citep{watson2023}. Crucially, this individual- and subgroup-level inconsistency directly conflicts with modern regulatory standards, most notably the EU AI Act\footnote{Regulation (EU) 2024/1689 of the European Parliament and of the Council laying down harmonised rules on artificial intelligence (AI Act).}, which emphasizes transparency and accountability in high-risk algorithmic assessments. To alleviate this instability, practitioners frequently resort to post-hoc calibration methods such as Platt Scaling \citep{platt1999probabilistic}, Isotonic Regression \citep{zadrozny2002}, or Beta Calibration \citep{kull2017beta}. The implicit, mechanical assumption underlying these workflows is that re-anchoring probability outputs to empirical frequencies is a benign adjustment that stabilizes model behavior.

In this paper, we challenge this foundational assumption by formalizing the \textit{over-calibration paradox}. We demonstrate that post-hoc calibration does not act as a universal remedy; instead, its capacity to harmonize predictions is strictly bounded by the native calibration quality of the underlying model space. While calibration serves as an effective regularizer in pipelines suffering from systematic miscalibration, forcing these same adjustments onto natively well-calibrated configurations produces the opposite effect. In sample-constrained portfolios or restricted subgroups, flexible post-hoc estimators overfit to hold-out noise, perturbing decision boundaries and increasing both expected calibration error and predictive multiplicity simultaneously. In high-stakes settings, this structural instability exposes a critical vulnerability to soft adversarial manipulation \citep{chou2026adversarial}, where unmonitored calibration choices can be strategically leveraged to steer contested individual decisions without degrading aggregate benchmark performance.

Two research questions guide our unified analysis: (i) How do prediction confidence and predictive multiplicity correlate under the Rashomon effect across diverse financial credit portfolios? (ii) What are the strict operational boundaries within which post-hoc calibration transitions from a regularizing agent into a destabilizing factor that triggers the over-calibration paradox? 

To address these questions, we used nine credit risk datasets to systematically extract diverse and structurally robust Rashomon sets. Our empirical findings yield three primary contributions:
\begin{enumerate}
    \item[(i)] We formalize and boundary-map the over-calibration paradox, identifying the empirical conditions under which post-hoc adjustments distort aligned decision boundaries and simultaneously inflate both expected calibration error and predictive multiplicity.
    \item[(ii)] We track the joint empirical distribution of model-level expected calibration error and individual obscurity across nine credit risk portfolios, separating the convergence regime where calibration suppresses multiplicity from the paradox regime where calibration introduces decision noise.
    \item[(iii)] We evaluate subpopulation disparities via non-parametric testing, showing that minority-class observations bear a disproportionate multiplicity burden. We demonstrate that unmonitored post-hoc calibration creates an unexamined decision margin, allowing candidate models to alter individual classifications while satisfying standard aggregate performance audits.
\end{enumerate}

\section{Related Work}

Auditing algorithmic decision systems beyond aggregate performance requires examining both the statistical reliability of probability estimates and the behavioral stability of competing models. This section surveys relevant literature across four interconnected dimensions: probabilistic calibration and finite-sample post-hoc adjustments, the emergence of prediction disagreement under the Rashomon effect, the structural divergence between probability alignment and decision boundary stability, and the strategic exploitation of model multiplicity in algorithmic auditing.

\subsection{Probabilistic Reliability and Post-hoc Adjustments}
\label{sec:rel_calib}
Classification calibration investigates the statistical agreement between a classifier's predicted confidence levels and its empirical outcome frequencies \citep{silva_filho_2023}. Early foundational work established that complex non-linear architectures frequently generate poorly calibrated, overconfident probability estimates \citep{niculescu2005,guo2017}, motivating a rich ecosystem of post-hoc methods including parametric sigmoidal mapping \citep{platt1999probabilistic}, non-parametric isotonic regression \citep{zadrozny2002}, and multiclass extensions \citep{widmann2019}. Beyond these canonical approaches, histogram binning \citep{zadrozny2001obtaining} and Bayesian Binning into Quantiles (BBQ) \citep{naeini2015} offer alternative non-parametric strategies that partition the probability space into discrete bins. Histogram binning offers simplicity and interpretability, while BBQ provides a principled Bayesian model-averaging approach that reduces sensitivity to bin boundary choices. We note that these methods are primarily designed for the binary classification setting. Extending them to multi-class classification introduces additional complexity, with approaches such as temperature scaling \citep{guo2017} and Dirichlet calibration \citep{kull2019dirichlet} addressing this challenge; their interaction with predictive multiplicity in multi-class settings remains an open research direction.

Recent literature highlights that the empirical success of these transformations depends heavily on data-centric traits. For instance, global metrics like Expected Calibration Error (ECE) often mask high demographic or subpopulation heterogeneity \citep{vaicenavicius2019,awe2024}. Furthermore, noise and data sparsity introduce severe estimation biases into post-hoc pipelines \citep{sinaga2025}. While these studies caution against uncritical calibration deployment, they primarily focus on how data properties degrade probability metrics, leaving the structural impact of calibration on individual decision boundaries largely unexamined.

\subsection{Prediction Disagreement under the Rashomon Effect}
\label{sec:rel_mult}
Predictive multiplicity captures a distinct dimension of algorithmic uncertainty: the non-uniqueness of individual predictions within an unnested model space \citep{marx2020}. Rooted in the Rashomon effect, where multiple distinct models achieve equivalent aggregate performance on the same dataset \citep{breiman2001}, predictive multiplicity exposes an under-explored form of algorithmic arbitrariness. Individuals with identical feature vectors can receive entirely conflicting classifications depending solely on which near-optimal model an auditor selects \citep{watson2023}.

Recent extensions have integrated these concerns into downstream socio-legal domains. Scholars have shown that imposing group-level fairness metrics can inadvertently expand alternative hypothesis spaces, exacerbating individual-level arbitrariness \citep{long2023}, a phenomenon that stakeholders routinely perceive as fundamentally unfair \citep{meyer2025}. Data-centric evaluations further confirm that preprocessing choices heavily dictate both the volume and behavioral diversity of the underlying Rashomon set \citep{stando2023,cavus_biecek_2025}. 

\subsection{Probability Calibration versus Boundary Stability}
\label{sec:intersection}
Despite these parallel advancements, the structural interaction between probability alignment and model multiplicity remains largely unaddressed. Emerging theoretical critique has begun to question whether traditional calibration metrics are sufficient to guarantee individual-level reliability \citep{machado2024}. Moreover, recent security audits demonstrate that uncertainty calibration is highly susceptible to adversarial perturbations \citep{decker2025}.

A common heuristic in machine learning assumes that probability calibration naturally stabilizes classifications near the decision threshold. However, aggregate frequency alignment does not imply local boundary stability. Calibrated probabilities reflect expected empirical proportions across broad intervals, whereas individual classifications depend on discrete thresholding. Indeed, theoretical analyses establish that even when competing models achieve close probability agreement almost everywhere, small residual discrepancies can still induce substantial losses and divergent downstream actions \citep{du2024reconciling}. Near a decision boundary ($\tau = 0.5$), minor shifts in estimated probabilities, driven by finite sample noise or non-parametric overfitting during post-hoc adjustment, can flip individual decisions without degrading overall calibration error. Consequently, boundary stability and probability calibration represent distinct statistical properties, motivating a unified analysis of their interaction across competing models.

\subsection{Strategic Exploitation of Model Multiplicity}
\label{sec:adversarial_multiplicity}
While predictive multiplicity has traditionally been framed as an opportunity to select models satisfying secondary socio-technical desiderata, such as group fairness or interpretability, without sacrificing aggregate accuracy \citep{semenova2022,black2022model}, emerging scholarship highlights that multiplicity also enables strategic or adversarial exploitation. When a large set of competing models achieves near-identical predictive performance, motivated actors can exploit the resulting flexibility to rationalize self-serving decisions under the guise of legitimate modeling choices \citep{ganesh2025systemizing,black2024dhacking}.

This concern connects to an established body of work on algorithmic rationalization and auditing evasion. Prior literature has documented phenomena such as \textit{fairwashing} \citep{aivodji2019fairwashing}, where developers select surrogate models or explanations that falsely appear nondiscriminatory, strategic metric selection or \textit{D-hacking} \citep{black2024dhacking,casper2024blackbox}, and \textit{x-hacking}, defined as the post-hoc cherry-picking of competing near-optimal models in automated pipelines to satisfy desired explanation narratives \citep{woznica2026equally}. Most recently, \citet{chou2026adversarial} formalize this landscape under the taxonomy of \textit{adversarial multiplicity}, distinguishing between \textit{hard manipulation}, which directly perturbs model weights or tree parameters, and \textit{soft manipulation}, which operates by cherry-picking among defensible modeling configurations, such as feature sets, hyperparameters, and post-hoc calibration functions.

While hard manipulation can be detected through white-box code verification, soft manipulation presents a distinct challenge: the resulting model is trained legitimately through standard pipelines and answers queries consistently, allowing it to pass both black-box audits and white-box code disclosures \citep{casper2024blackbox}. In this paper, we bridge this emerging governance literature with probabilistic reliability, demonstrating that the \textit{over-calibration paradox} represents an unmonitored operational surface where routine post-hoc adjustments introduce the exact prediction volatility exploited by soft manipulation.

\section{Methodology}

\subsection{Predictive Multiplicity}

The phenomenon of \textit{predictive multiplicity} refers to the situation where multiple models, despite having near-identical prediction performance, yield substantially different predictions for the same individual instances \citep{marx2020}. This is fundamentally driven by the \textit{Rashomon effect} \citep{breiman2001}.

Formally, let $\mathcal{F}$ be a hypothesis class and $L(f, \mathcal{D})$ denote the loss of model $f$ on dataset $\mathcal{D}$. For a given tolerance $\epsilon > 0$, the \textit{Rashomon set} $\mathcal{R}$ is formalized using the Rashomon ratio framework \citep{semenova2022}:
\begin{equation}
    \mathcal{R}(\epsilon) = \{ f \in \mathcal{F} : L(f, \mathcal{D}) \leq L(f_{best}, \mathcal{D}) + \epsilon \}
\end{equation}

\subsubsection{Obscurity}
\textit{Obscurity} measures the mean disagreement rate between near-optimal models in the Rashomon set and the empirically best-performing baseline classifier $f_{best}$ \citep{cavus_biecek_2025}:
\begin{equation}
    \gamma_\epsilon(x) = \frac{1}{|\mathcal{R}| - 1} \sum_{f \in \mathcal{R} \setminus \{f_{best}\}} \mathbb{I}(f(x) \neq f_{best}(x))
\end{equation}
\noindent where $\mathbb{I}(\cdot)$ is the indicator function. Anchoring the disagreement rate to $f_{best}$ directly reflects realistic auditing and compliance scenarios where contested decisions are evaluated against an incumbent or deployed production model. From a security standpoint, obscurity captures the fine-grained fragility of each individual's classification; an individual instance $x$ is susceptible to adversarial cherry-picking under model multiplicity \citep{chou2026adversarial} whenever $\gamma_\epsilon(x) > 0$, with higher obscurity reflecting severe decision arbitrariness across defensible pipelines.

\subsubsection{Classification Thresholding}
A standard classification threshold of $\tau = 0.5$ is applied across all models to convert predicted probabilities into discrete decision labels $f(x) \in \{0, 1\}$:
\begin{equation}
    f(x) = \mathbb{I}(\hat{p}(x) > 0.5).
\end{equation}

\subsection{Post-hoc Calibration Methods}

Post-hoc calibration seeks a transformation $\hat{p}_{\mathrm{cal}}(x) = g(\hat{p}(x))$ learned on a hold-out calibration dataset \citep{zadrozny2002,niculescu2005}. Strictly monotone calibration functions, such as Platt Scaling and Beta Calibration, preserve the ranking of baseline predictions; in contrast, non-parametric methods such as Isotonic Regression can disrupt rankings when fitted on sample-constrained calibration sets with ties. Calibration transformations are fitted independently to each model in the Rashomon set using the calibration partition, ensuring that post-hoc adjustments do not artificially constrain model diversity by construction.

\subsubsection{Platt Scaling} is a parametric method \citep{platt1999probabilistic} that models calibrated probability as a logistic transformation:
\begin{equation}
    \hat{p}_{\mathrm{cal}}(x) = \frac{1}{1 + \exp\left( A \hat{p}(x) + B \right)},
\end{equation}
\noindent where $A$ and $B$ are estimated by minimizing the negative log-likelihood on the calibration set.

\subsubsection{Isotonic Regression} is a non-parametric method that fits a monotonically non-decreasing function by minimizing
\begin{equation}
    \sum_{i=1}^m \left( y_i - g(\hat{p}(x_i)) \right)^2,
\end{equation}
\noindent subject to monotonicity constraints \citep{zadrozny2002}. Its flexibility may lead to overfitting when the calibration set is small.

\subsubsection{Beta Calibration} generalizes Platt scaling using a beta distribution transformation \citep{kull2017beta}:
\begin{equation}
    \hat{\beta}_{\mathrm{cal}}(x) = \frac{1}{1 + \exp\left( -a \ln(\hat{p}(x)) + b \ln(1 - \hat{p}(x)) - c \right)},
\end{equation}
\noindent where $a, b, c \in \mathbb{R}$ are estimated by optimizing the log-likelihood, subject to $a \ge 0$ and $b \ge 0$ to ensure monotonicity.

\subsection{Confidence}
\begin{equation}
    \text{conf}_f(x) = \max \Big( \hat{p}_f(x), 1 - \hat{p}_f(x) \Big),
\end{equation}
\noindent where $\hat{p}_f(x) = P(Y=1 \mid x)$. This bounds confidence within $[0.5, 1]$. Mean Rashomon-set confidence is:
\begin{equation}
    \text{conf}_{\mathcal{R}}(x) = \frac{1}{|\mathcal{R}|} \sum_{f \in \mathcal{R}} \text{conf}_f(x).
\end{equation}

\subsection{Expected Calibration Error}
We employ the Expected Calibration Error (ECE) \citep{guo2017} to evaluate probabilistic reliability before and after post-hoc calibration. The $[0,1]$ probability interval is partitioned into $M$ equally spaced bins $B_m = (\frac{m-1}{M}, \frac{m}{M}]$. To ensure unbiased assessment, calibration error is evaluated strictly on the held-out test partition. The aggregate ECE is defined as:
\begin{equation}
    \text{ECE} = \sum_{m=1}^{M} \frac{|B_m|}{N} \left| \mathrm{accuracy}(B_m) - \mathrm{conf}(B_m) \right|
\end{equation}
\noindent where $N$ is the total number of test observations and we fix $M = 10$ uniformly spaced bins. Because ECE aggregates miscalibration into a single scalar it may mask within-bin heterogeneity; we therefore report it jointly with obscurity to capture both aggregate and individual-level instability.

\subsection{The Over-Calibration Paradox}
\label{sec:paradox}
We formalize the \textit{over-calibration paradox} as the empirical regime in which applying post-hoc probability calibration to a natively well-calibrated classifier configuration simultaneously increases both expected calibration error and predictive multiplicity relative to the uncalibrated baseline. Formally, for a given dataset $\mathcal{D}$, reference model $f_{best}$, and Rashomon set $\mathcal{R}$, the paradox occurs when:
\begin{equation}
    \mathrm{ECE}(g(f)) > \mathrm{ECE}(f)
\quad \text{and} \quad
\frac{1}{|\mathcal{D}|} \sum_{x \in \mathcal{D}} \gamma_\epsilon^{\mathrm{cal}}(x)
> \frac{1}{|\mathcal{D}|} \sum_{x \in \mathcal{D}} \gamma_\epsilon(x)
\end{equation}
\noindent for post-hoc transformations $g \in$ \{Platt, Isotonic, Beta\}. We conjecture that this paradox arises when: (i) baseline models already occupy a low-ECE region, and (ii) the calibration hold-out is too small to support reliable estimation, causing flexible non-parametric estimators to overfit to sample noise and distort well-aligned decision boundaries.

\section{Experiments}

The experimental framework, outlined in Figure~\ref{fig:workflow_overview}, evaluates the effect of post-hoc calibration methods on predictive multiplicity. We leverage the AutoML tool h2o \citep{h2o} to generate near-optimal models and focus primarily on the \textit{obscurity} metric \citep{cavus_biecek_2024}.

\begin{figure}[h]
    \centering
    \resizebox{\linewidth}{!}{%
    \begin{tikzpicture}[
        node distance=9mm and 8mm,
        box/.style={rectangle, rounded corners=2.8mm, draw=black!75, line width=0.9pt, align=center, minimum height=13mm, text width=4.3cm},
        arrow/.style={-Latex, line width=1.05pt, draw=black!80}
    ]
        \node[box, fill=teal!18] (split) {\textbf{Data Split}\\Train (60\%) \\ Calibration (20\%) \\ Test (20\%)};
        \node[box, fill=cyan!20, right=of split] (automl) {\textbf{AutoML Model Set}\\h2o AutoML (GBM, RF,\\DNN, GLM; $n=20$)};
        \node[box, fill=violet!17, right=of automl] (rashomon) {\textbf{Rashomon Set}\\top-$\epsilon$ models\\($\epsilon=0.05$, AUC)};
        \node[box, fill=green!18, below=of rashomon] (evalraw) {\textbf{Evaluation (Raw)}\\Obscurity \\  ECE \\ Confidence};
        \node[box, fill=orange!20, right=of rashomon] (calib) {\textbf{Post-hoc Calibration}\\Platt Scaling \\ Isotonic Regression \\ Beta Calibration};
        \node[box, fill=green!18, right=of evalraw] (evalcal) {\textbf{Evaluation (Calibrated)}\\Obscurity \\ ECE \\ Confidence};
        \node[box, fill=pink!20, below=7mm of evalcal, xshift=-26.5mm] (compare) {\textbf{Comparison}\\metrics\\+ statistical tests};
        \draw[arrow] (split) -- (automl);
        \draw[arrow] (automl) -- (rashomon);
        \draw[arrow] (rashomon) -- (calib);
        \draw[arrow] (rashomon) -- (evalraw);
        \draw[arrow] (calib) -- (evalcal);
        \draw[arrow] (evalraw.south) |- (compare.west);
        \draw[arrow] (evalcal.south) |- (compare.east);
    \end{tikzpicture}%
    }
    \caption{Workflow of the experimental methodology. The data is split 60/20/20 into train, calibration, and test sets. h2o AutoML trains up to 20 classifiers spanning Gradient Boosting Machines (XGBoost and H2O GBM with varied depth and learn-rate), Random Forests (Distributed RF, 50 trees), Deep Neural Networks (fully-connected with ReLU and dropout), and Generalized Linear Models (elastic-net logistic regression). The Rashomon set retains all models within $\epsilon=0.05$ AUC of the top model. Each model is then evaluated raw and after each of three post-hoc calibration functions; all metrics (obscurity, ECE, confidence) are computed on the held-out test set and compared via non-parametric statistical tests.}
    \label{fig:workflow_overview}
\end{figure}

\subsection{Setup}

Experiments are conducted using the scalable AutoML engine \texttt{h2o} (v3.44.0.3) \citep{h2o}. To ensure structural diversity across the candidate hypothesis space $\mathcal{F}$, the automated search protocol trains up to 20 candidate models per portfolio spanning four distinct algorithm families: gradient boosted decision trees and XGBoost, distributed random forests, fully-connected deep neural networks, and elastic-net regularized generalized linear models. The exploration search is conducted under fixed random seeds to ensure full deterministic reproducibility. Following standard predictive multiplicity methodology \citep{marx2020}, the Rashomon set $\mathcal{R}$ retains all models whose validation AUC falls within a 5\% relative tolerance of the top-performing classifier $f_{best}$:
\begin{equation}
    \mathcal{R}(\epsilon) = \left\{ f \in \mathcal{F} : \mathrm{AUC}(f) \ge \mathrm{AUC}(f_{best}) \times (1 - \epsilon) \right\}, \quad \epsilon = 0.05.
\end{equation}

Each dataset is partitioned into training (60\%), calibration (20\%), and test (20\%) sets. Post-hoc calibration functions are fitted on the calibration split, while prediction confidence, expected calibration error, and obscurity scores are evaluated strictly on the held-out test partition. All replication code and benchmark data are hosted in our public repository at \url{https://github.com/mcavs/overcalibration-paradox}.

\subsection{Data}

We evaluate predictive multiplicity using nine publicly available credit risk scoring datasets detailed in Table~\ref{tab:datasets}. Observations range from 1,000 to 251,503 and feature counts from 11 to 65. The imbalance ratio ($n_{\mathrm{majority}} / n_{\mathrm{minority}}$) ranges from 2.3 to 20.2. Credit scoring datasets are selected because (i) they exhibit a wide range of naturally occurring imbalance ratios enabling systematic examination of calibration under data sparsity, and (ii) they are publicly available benchmarks that support reproducibility. The Rashomon ratio ranges from 0.45 to 1.

\begin{table}[h]
\centering
\caption{Summary of credit risk datasets and their respective Rashomon set properties.}
\label{tab:datasets}
\setlength{\tabcolsep}{5pt}
\resizebox{\textwidth}{!}{
\begin{tabular}{lrrrrr}
\toprule
Dataset & \#Observation & \#Variable & \shortstack[r]{Imbalance \\ Ratio} & \shortstack[r]{Rashomon \\ Ratio} & \shortstack[r]{Best \\ Model \\ AUC} \\ 
\midrule
\texttt{AER\_credit\_card\_data} & 1,319   & 12 & 3.426   & 1       & 0.9946 \\
\texttt{bank\_marketing}         & 45,211  & 17 & 7.548   & 1       & 0.8831 \\
\texttt{german\_credit}          & 1,000   & 21 & 2.333   & 0.80    & 0.7693 \\
\texttt{give\_me\_credit}        & 251,503 & 11 & 13.961  & 1       & 0.8561 \\
\texttt{hmeq}                    & 5,960   & 13 & 4.012   & 0.45    & 0.8923 \\
\texttt{loan\_data}              & 1,225   & 15 & 2.792   & 0.75    & 0.6158 \\
\texttt{poland\_year3}           & 10,503  & 65 & 20.218  & 0.60    & 0.8649 \\
\texttt{poland\_year5}           & 5,910   & 65 & 13.414  & 0.65    & 0.9257 \\
\texttt{taiwan\_credit}          & 30,000  & 24 & 3.520   & 1       & 0.7769 \\ 
\bottomrule
\end{tabular}}
\end{table}

\section{Results}

Our empirical evaluation yields a multi-layered understanding of how statistical probability alignment interacts with algorithmic prediction stability under the Rashomon effect.

\subsection{Baseline Multiplicity, Prediction Confidence, and Subpopulation Disparities}

\begin{figure}[t]
    \centering
    \includegraphics[width=\textwidth]{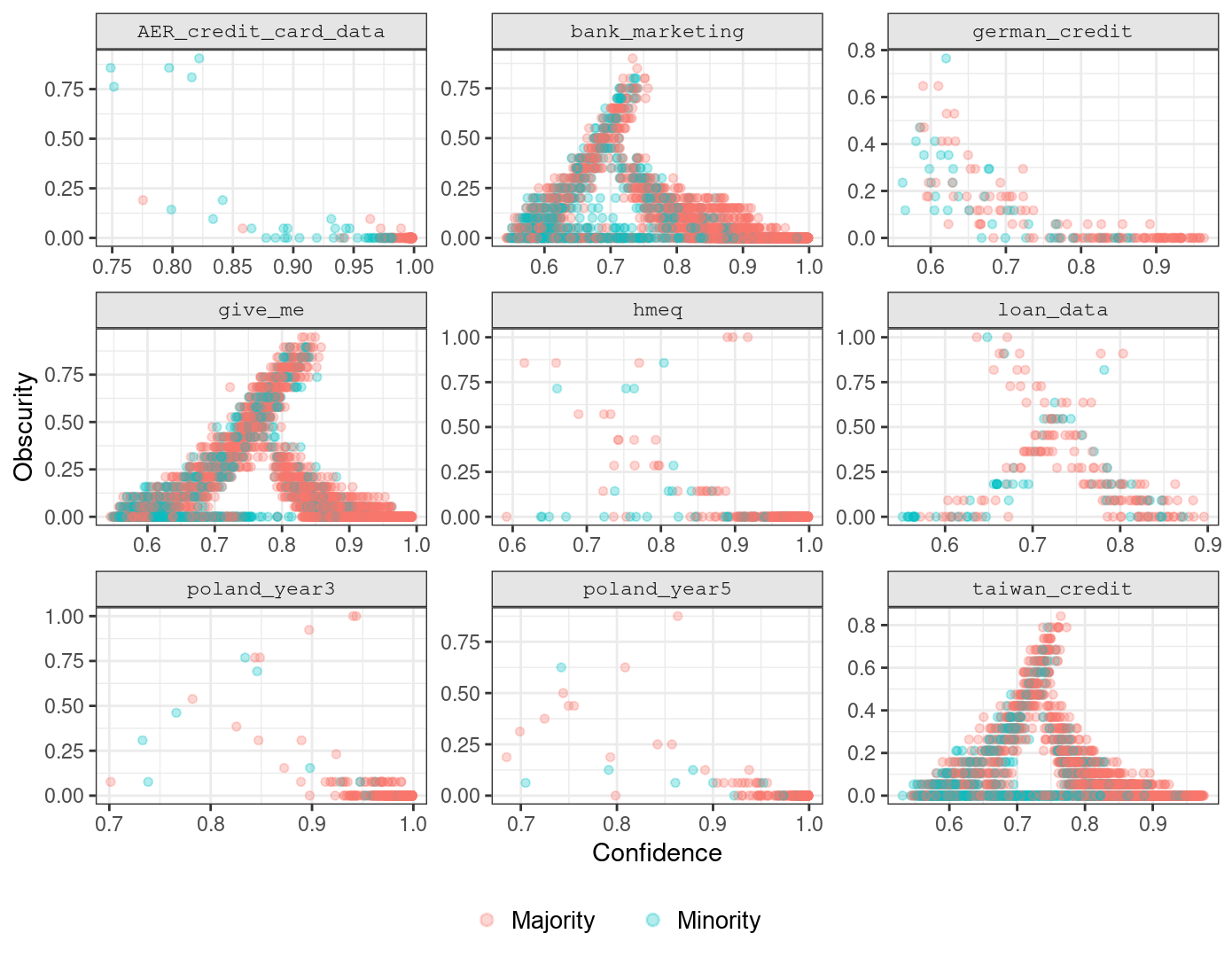}
    \caption{Relationship between Prediction Confidence and Obscurity across nine credit scoring datasets for majority-minority class distributions.}
    \label{fig:rashomon_analysis}
\end{figure}

Before evaluating post-hoc interventions, we map the structural baseline of uncalibrated Rashomon sets. Our empirical analysis confirms a robust, systemic inverse correlation between average model confidence ($\text{conf}_{\mathcal{R}}(x)$) and individual predictive multiplicity ($\gamma_\epsilon(x)$) across all evaluated portfolios as illustrated in Figure~\ref{fig:rashomon_analysis}. In high-confidence regions ($\text{conf}_{\mathcal{R}}(x) > 0.90$), near-optimal models converge toward consensus, resulting in near-zero obscurity. As predictions approach the decision threshold ($\tau = 0.5$), confidence plummets and individual obscurity spikes to disagreement rates between $50\%$ and $80\%$. This relationship indicates that relying on a single top-performing model obscures a significant range of valid, conflicting alternative predictions.

Our non-parametric statistical testing proves that baseline instability is not uniformly distributed but imposes a disproportionate level of predictive multiplicity on minority sub-populations. Wilcoxon rank-sum tests \citep{wilcoxon1945} reveal that minority class observations exhibit significantly higher baseline obscurity ($W = 36,894,926, p < .001$) and lower average confidence ($W = 89,252,968, p < .001$). A chi-squared test confirms minority observations are significantly more vulnerable to critical decision flips ($\chi^2(1, N = 10,503) = 1,885.8, p < .001$). This structural ambiguity establishes the motivation for deploying post-hoc calibration as a potential harmonizing regularizer.

\subsection{The Over-Calibration Paradox and Post-hoc Calibration Regimes}

\begin{figure}[t]
    \centering
    \includegraphics[width=\textwidth]{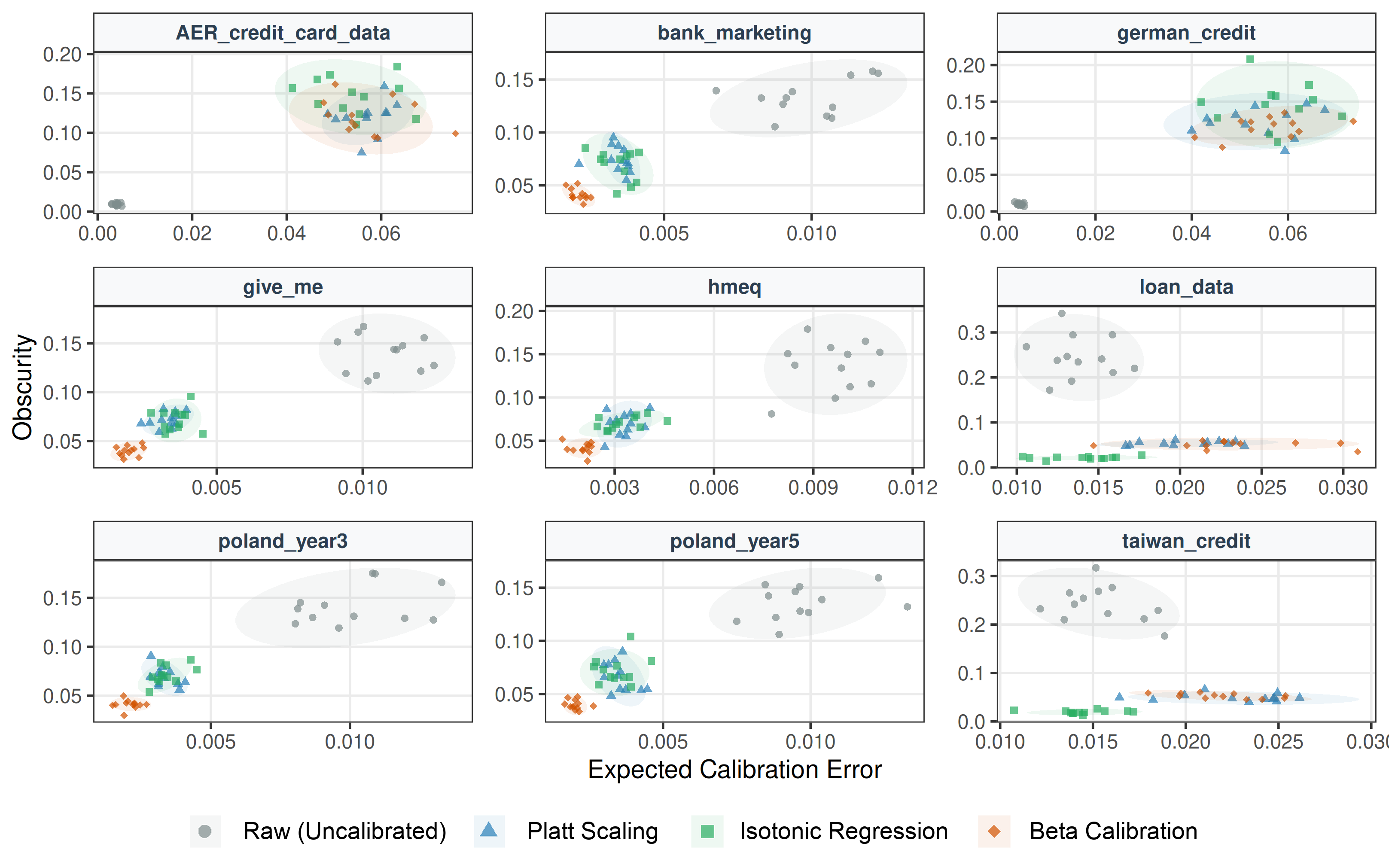}
    \caption{Empirical distribution of \textit{expected calibration error} versus \textit{mean obscurity} across all nine credit scoring benchmark datasets. Each point represents a near-optimal model within the Rashomon set. The shaded areas denote the $90\%$ confidence ellipses for each method.}
    \label{fig:ece}
\end{figure}

To track how post-hoc functions alter this baseline instability, we map the joint empirical distribution of model-level ECE and individual-level mean obscurity depicted in Figure~\ref{fig:ece}. Our analysis demonstrates clear empirical limits to the assumption that post-hoc calibration is a universally benign refinement step. Instead, the interaction reveals two strictly bounded, opposing behavioral regimes:

\begin{enumerate}
    \item \textbf{The Alignment and Convergence Regime:} In datasets with severe native probabilistic misalignment (\texttt{bank\_marketing}, \texttt{give\_me}, \texttt{hmeq}, \texttt{poland\_year3}, \texttt{poland\_year5}), where baseline models exhibit $\text{ECE} > 0.010$ and substantial initial obscurity, post-hoc interventions act as effective regularizers. In this regime, Platt Scaling, Isotonic Regression, and Beta Calibration successfully compress the alternative hypothesis space, reducing minority class mean obscurity from $0.138 \pm 0.035$ to below $0.090$, as reported in Table~\ref{tab:combined_results}. Here, empirical probabilities are brought closer to empirical reality, and conflicting classifications among near-optimal models are largely suppressed.
  
    \item \textbf{The Over-Calibration Paradox Regime:} A notable divergence occurs in the \texttt{AER\_credit\_card\_data} and \texttt{german\_credit} portfolios. Both represent sample-constrained domains (1,319 and 1,000 observations, respectively) whose uncalibrated baseline classifiers are already inherently well-calibrated. Enforcing post-hoc probability adjustments on these stable structures triggers a simultaneous explosion in both expected calibration error and predictive multiplicity. As depicted in Figure~\ref{fig:ece}, while larger portfolios migrate toward lower ECE and reduced obscurity under calibration, models for \texttt{german\_credit} shift sharply into high-multiplicity, high-ECE space. 
    
    This structural inversion directly reflects the finite-sample dynamics of post-hoc estimators: with held-out calibration partitions of only 200 to 264 instances, flexible non-parametric step functions, such as Isotonic Regression, overfit to empirical local noise. Rather than refining the underlying probabilities, these transformations introduce artificial step-discontinuities across the scoring distribution, perturbing individual margins near the decision boundary ($\tau = 0.5$) and generating spurious prediction disagreements across the Rashomon set. 
  
    Remarkably, this empirical phenomenon is independently corroborated by concurrent findings in adversarial machine learning: \citet{chou2026adversarial} evaluate model multiplicity across four fairness benchmarks and observe that while multi-metric auditing typically restricts manipulation reach, on the \texttt{german\_credit} dataset, introducing calibration checks paradoxically increases adversarial manipulation reach by $+9.5$ percentage points. The fact that two independent investigations, one conducted through the lens of algorithmic obscurity and our own conducted through the lens of post-hoc calibration distortion, arrive at the exact same destabilization anomaly in \texttt{german\_credit} confirms that the over-calibration paradox is not an empirical artifact, but a fundamental structural vulnerability inherent to constrained tabular regimes.
\end{enumerate}

\begin{table}[t]
\centering
\setlength{\tabcolsep}{4pt}
\caption{Summary and Stratified Post-hoc Dunn Test Results Comparing Calibration Methods Against the Baseline for \textit{Obscurity} and \textit{Confidence}.}
\label{tab:combined_results}
\resizebox{\textwidth}{!}{
\begin{tabular}{ll crr crr}
\toprule
& & \multicolumn{3}{c}{Obscurity ($\downarrow$)} & \multicolumn{3}{c}{Confidence ($\uparrow$)} \\
\cmidrule(lr){3-5} \cmidrule(lr){6-8}
Group & Method & Mean $\pm$ SE & $Z$ & $p_{adj}$ & Mean $\pm$ SE & $Z$ & $p_{adj}$ \\
\midrule
\multirow{4}{*}{Majority} & Uncalibrated (Raw) & $0.062 \pm 0.021$ & --- & --- & \cellcolor{green!20}$0.912 \pm 0.015$ & --- & --- \\
                  & Platt Scaling & \cellcolor{green!20}$0.021 \pm 0.008$ & -41.3 & $<.001$ & $0.901 \pm 0.012$ & -26.6 & $<.001$ \\
                  & Isotonic Regression & $0.024 \pm 0.009$ & -41.1 & $<.001$ & $0.895 \pm 0.014$ & 19.6  & $<.001$ \\
                  & Beta Calibration & $0.018 \pm 0.005$ & -38.4 & $<.001$ & $0.908 \pm 0.011$ & 14.2  & $<.001$ \\
\midrule
\multirow{4}{*}{Minority} & Uncalibrated (Raw) & $0.138 \pm 0.035$ & --- & --- & $0.808 \pm 0.028$ & --- & --- \\
                  & Platt Scaling & \cellcolor{green!20}$0.084 \pm 0.022$ & -5.62 & $<.001$ & \cellcolor{green!20}$0.835 \pm 0.021$ & 13.0  & $<.001$ \\
                  & Isotonic Regression & $0.098 \pm 0.025$ & -4.19 & $<.001$ & $0.811 \pm 0.024$ & -1.26 & $.206$ \\
                  & Beta Calibration & $0.092 \pm 0.020$ & -4.91 & $<.001$ & $0.814 \pm 0.022$ & 1.04  & $.298$ \\
\bottomrule
\addlinespace
\multicolumn{8}{p{13.2cm}}{\small \textit{Note.} The $Z$-statistic and adjusted $p_{adj}$-values denote post-hoc pairwise statistical comparisons against the Uncalibrated (Raw) baseline. All $p$-values are adjusted using the Bonferroni correction \citep{bonferroni1936} ($N = 133,852$).}
\end{tabular}}
\end{table}

To isolate the technical drivers, we examine the stratified Dunn test results \citep{dunn1964} presented in Table~\ref{tab:combined_results}. Isotonic Regression and Beta Calibration fail to significantly refine minority confidence levels ($p_{adj} > .05$). Because these flexible estimators overfit to sampling noise within restricted calibration splits (typically $200$--$260$ observations in smaller portfolios), they introduce artifactual boundary noise into inherently well-aligned classifiers. Platt Scaling remains the only method to significantly improve minority confidence scores ($Z = 13.0, p < .001$), suggesting that parametric restrictions provide a necessary defense against overfitting under data scarcity.

\section{Discussion and Conclusion}

In this study, we bridged two foundational pillars of trustworthy AI: statistical classification calibration and predictive multiplicity. Our analysis reveals that these two dimensions are closely connected.

Regarding our first research question, we confirm a strong, non-linear inverse correlation between prediction confidence and predictive multiplicity. Observations near the decision boundary suffer from intense prediction volatility, and our non-parametric testing validates that these structural ambiguities impose a disproportionate level of predictive multiplicity on minority subpopulations. In credit scoring, this represents a systemic fairness issue: higher-risk applicants are significantly more likely to experience \textit{luck-of-the-draw} decisions depending entirely on an arbitrary choice among multiple equally accurate models. This individual-level arbitrariness directly collides with modern regulatory frameworks such as the EU AI Act.

In addressing our second research question, we formalize the \textit{over-calibration paradox} and expose the strict operational boundaries of post-hoc interventions. We demonstrate that calibration acts as a powerful regularizer \textit{only} when rectifying a genuine native probability misalignment. When deployed needlessly on natively well-calibrated configurations, or when subpopulation data sparsity limits sample density in the calibration hold-out, flexible post-hoc procedures introduce severe estimation noise. This overfitting behavior distorts aligned decision thresholds and artifactually inflates both calibration error and predictive multiplicity simultaneously.

These insights are particularly critical for security-sensitive deployment pipelines. We re-conceptualize the baseline Rashomon set and calibration pipeline as an unmonitored decision margin. Rather than relying on standard probability adjustments that may inadvertently inflate instability, alternative governance directions should explore direct Rashomon-set reconciliation and local patching techniques designed to resolve prediction inconsistencies across competing models \citep{haghighat2026resolving}. In adversarial settings, an unmonitored calibration pipeline could be selectively leveraged to alter individual outcomes for specific demographic groups while satisfying global performance metrics.

\paragraph{Limitations.}
\label{sec:limitations}

Several limitations of the present study should be acknowledged. First, our empirical evaluation is confined to nine credit scoring datasets and the binary classification setting. Generalization to multi-class problems, which require different calibration strategies such as Dirichlet calibration \citep{kull2019dirichlet} and distinct multiplicity metrics, is left for future work. Second, we restricted our analysis to three calibration methods. Other approaches such as histogram binning \citep{zadrozny2001obtaining} and BBQ \citep{naeini2015} may exhibit different over-calibration behavior and warrant dedicated investigation. Third, our Rashomon sets are constructed using a single AutoML framework (\texttt{h2o}). As demonstrated in recent AutoML benchmarking studies, different automated frameworks explore hypothesis spaces dissimilarly; for example, AutoGluon tends to generate broader and structurally more diverse Rashomon sets than \texttt{h2o}, which can directly influence the baseline volume of predictive multiplicity and the resulting vulnerability to soft manipulation \citep{woznica2026equally}. Evaluating the over-calibration paradox across alternative AutoML frameworks and tabular foundation models represents an important avenue for future research. Fourth, while we conjecture the paradox is driven by insufficient calibration sample size, a formal theoretical characterization of the critical sample size threshold remains an open problem.

In conclusion, our work demonstrates that calibration quality and model stability cannot be decoupled. For practitioners deploying trustworthy AI, our findings offer an actionable warning: probability adjustments must be rigorously audited together with alternative hypothesis spaces, as unaudited refinement routines actively jeopardize the individual-level safety and robustness of critical decision systems.

\begin{credits}
\subsubsection{\discintname}
The authors have no competing interests to declare that are relevant to the content of this article.
\end{credits}

\end{document}